\documentclass[runningheads]{llncs}
\usepackage[T1]{fontenc}
\usepackage{graphicx}
\usepackage{booktabs}
\usepackage{float}
\usepackage{multirow}
\usepackage{seqsplit}
\usepackage{amsmath, amssymb}

\usepackage{color}
\usepackage{arydshln}
\begin{document}
\title{TopicVD: A Topic-Based Dataset of Video-Guided Multimodal Machine Translation for Documentaries}
%
%
\author{Jinze Lv\inst{1} \and
Jian Chen\inst{1} \and
Zi Long\inst{2}\thanks{Corresponding author} \and
Xianghua Fu\inst{2} \and
Yin Chen\inst{2}}
\authorrunning{J. Lv et al.}
%
\institute{College of Application and Technology, Shenzhen University, China \and
College of Big Data and Internet, Shenzhen Technology University, China
\email{\{longzi,fuxianghua, chenyin\}@sztu.edu.cn}}
\maketitle              
\begin{abstract}
Most existing multimodal machine translation (MMT) datasets are predominantly composed of static images or short video clips, lacking extensive video data across diverse domains and topics. As a result, they fail to meet the demands of real-world MMT tasks, such as documentary translation.
In this study, we developed TopicVD, a topic-based dataset for video-supported multimodal machine translation of documentaries, aiming to advance research in this field.
We collected video-subtitle pairs from documentaries and categorized them into eight topics, such as economy and nature, to facilitate research on domain adaptation in video-guided MMT. 
Additionally, we preserved their contextual information to support research on leveraging the global context of documentaries in video-guided MMT.
To better capture the shared semantics between text and video, we propose an MMT model based on a cross-modal bidirectional attention module.
Extensive experiments on the TopicVD dataset demonstrate that visual information consistently improves the performance of the NMT model in documentary translation. 
However, the MMT model's performance significantly declines in out-of-domain scenarios, highlighting the need for effective domain adaptation methods.
Additionally, experiments demonstrate that global context can effectively improve translation performance.
Dataset and our implementations are available at
\url{https://github.com/JinzeLv/TopicVD}


\end{abstract}

\section{Introduction}
\label{sec:Intro}

Machine translation has evolved significantly over the past few decades, from rule-based and statistical methods to deep learning-driven neural machine translation (NMT)~\cite{brown1990statistical,brown1993mathematics}.
In recent years, some scholars have proposed using visual data as auxiliary information to aid translation, such as Multi30K~\cite{elliott2016multi30k,plummer2015flickr30k,lin2014microsoft,hu2023large}. 
Although Multi30K has demonstrated its potential, the dataset primarily focuses on simple, short bilingual and image-text pairing tasks, with one-to-one bilingual mappings. This limitation results in insufficient contextual information in the source text, preventing it from accurately reflecting the translation scenario~\cite{caglayan2019probing,tang2022multimodal}.

To address complex translation scenarios and meet real-world needs, some scholars have proposed video-guided machine translation (VMT)~\cite{wang2019vatex}. 
Due to the limited availability of VMT datasets, researchers have focused on expanding datasets, such as How2~\cite{sanabria2018how2} and VATEX~\cite{wang2019vatex}.
However, previous VMT datasets have prioritized data quantity over quality, with many video clips consisting of subtitle descriptions and featuring discontinuous or excessively short durations. 
In recent years, the emergence of VISA~\cite{li2022visa}, EVA~\cite{li2023video}, BigVideo~\cite{kang2023bigvideo}, and TAIL~\cite{teramen2024english} has enhanced VMT datasets, enabling them to contain millions of data points and feature long videos ranging from ten to tens of minutes.
However, these datasets are not categorized by video topic; instead, they aggregate all data, which makes it difficult to analyze the impact of in-domain versus out-of-domain data on VMT.
Moreover, these datasets primarily focus on the correspondence between video clips and text, while neglecting the contextual information of video-text pairs, thereby limiting their applicability to VMT research that relies on global context.

Therefore, to support the above VMT research, a multimodal translation dataset is required that includes topic annotations for each data instance and retains the contextual structure of video clips within their corresponding videos.
In this study, we selected documentaries as the foundation for our translation dataset and constructed a topic-based corpus, TopicVD, to support research in video-supported multimodal machine translation.
TopicVD consists of 256 documentaries across 8 topics, totaling 285 hours in duration, and 122,930 Chinese-English parallel subtitle pairs, with a data size ranging from 3K to 30K for each topic.
We highlight the key features of TopicVD as follows:
a) The data in TopicVD are categorized into eight topics: Economy, Food, History, Figure, Military, Nagure, Social and Technology.
b) To support VMT research that utilizes global context, TopicVD preserves the contextual information of each video-subtitle pair.
To better model the shared semantics between text and video, we propose an MMT model based on a cross-modal bidirectional attention module. Extensive experiments on the TopicVD dataset demonstrate that visual information consistently improves the performance of the NMT model in documentary translation.
Further experimental analysis validated that the effectiveness of VMT for documentaries is significantly influenced by in-domain and out-of-domain factors, highlighting the need for effective domain adaptation methods. 
Additionally, experiments demonstrate that global information can significantly improve translation performance.

To summarize, our contributions are primarily in the following three areas:
\begin{itemize}
  \setlength{\itemsep}{0pt}
  \item[(1)] We constructed a topic-based dataset for video-supported multimodal machine translation of documentaries.
  \item[(2)] To better model the shared semantics between text and video, we propose an MMT model based on a cross-modal bidirectional attention module.
  \item[(3)] We validated the effectiveness of the proposed model on TopicVD and further investigated the impact of in-domain and out-of-domain data, as well as contextual information, on VMT for documentaries.
\end{itemize}

\section{Related Work}
\label{sec:rel}
\subsection{Image-guided Machine Translation}
\label{sec:imt}
To address the issues caused by the lack of context in traditional text translation, scholars have initially used images as additional inputs for multimodal machine translation~\cite{hitschler2016multimodal}. They proposed achieving higher efficiency than neural machine translation by using a seq-to-seq model with strong image features~\cite{elliott2017imagination,yao2020multimodal,lin2020dynamic,yin2020novel,su2021multi,li2022vision,lin2020dynamic,zhu2022beyond,lan2023exploring}.
One of the most widely used datasets is Multi30K~\cite{elliott2016multi30k,elliott2017findings}.
Multi30K consists of 30K images with relatively simple sentence structures and lacks complex grammatical features such as long sentences and nested clauses, making it unsuitable for translation tasks in complex contexts.
Some scholars have criticized Multi30K, arguing that its images provide insufficient information~\cite{wu2021good,li2021vision}.
Tang et al.~\cite{tang2022multimodal} explored the use of search engines to retrieve images as auxiliary information for MMT, achieving certain improvements. However, this approach remains limited to static images, which do not fully align with the real-world scenarios of MMT.
In most real-world MMT scenarios, text translation typically relies on video data.
Unlike images, which provide static visual cues, videos offer richer visual information by capturing temporal dynamics, including actions and scene transitions.
Therefore, we aim to construct a video-subtitle dataset to support research on video-guided machine translation for documentaries.

\subsection{Video-guided Machine Translation}
\label{sec:vmt}

To address complex translation scenarios and meet real-world needs, some scholars have proposed video-guided machine translation (VMT)~\cite{wang2019vatex}.
In the field of VMT, How2~\cite{sanabria2018how2} and VATEX~\cite{wang2019vatex} were the first datasets introduced for VMT.
VATEX contains 129K bilingual subtitle pairs and their corresponding video clips. However, studies have shown that translations can often be performed solely based on VATEX subtitle pairs, with video information frequently being ignored~\cite{yang2022videos}.
How2 collects tutorial videos from the YouTube platform, with an average video length of 90 seconds, and a total of 186K bilingual subtitle pairs.
Moreover, VISA~\cite{li2022visa} constructs a 40K VMT dataset to emphasize the importance of ambiguities, with each video lasting 10 seconds.
TAIL~\cite{teramen2024english} builds a 70K video-subtitle translation dataset for classroom tutorial videos.
BigVideo~\cite{kang2023bigvideo} and EVA~\cite{li2023video} expand the size and diversity of VMT datasets by releasing a dataset containing one million video-subtitle translation pairs.
The small data sizes of VISA and TAIL can lead to insufficient model training, making it difficult to demonstrate that additional visual information improves translation quality when compared to millions of text-based machine translation datasets.
VATEX, How2, and BigVideo have sufficient data sizes, but their data primarily focus on short-video platforms like YouTube~\footnote{
\url{https://www.youtube.com}
} and Xigua Video~\footnote{
\url{https://www.ixigua.com}
}, making it challenging to extract key information for complex tasks and scenarios.
Although EVA has a large-scale dataset with millions of video samples, the average length of each subtitle text in the EVA dataset is only 7.34 English words. We argue that movies and TV dramas often contain dialects, colloquial expressions, and complex emotional nuances, which are difficult to directly transfer to formal translation scenarios.
%
However, these datasets are not categorized by video topic; instead, they aggregate all data, making it difficult to analyze the impact of in-domain versus out-of-domain data on VMT.
In this study, we constructed a topic-based dataset for video-supported multimodal machine translation of documentaries to advance research in documentary translation using multimodal machine translation.
%
%
Table~\ref{tab:datasets} provides a detailed comparison of existing VMT datasets.

\begin{table}[t]
    \centering
    \caption{Comparison of Multimodal Translation Datasets}
    \label{tab:datasets}
    \begin{tabular}{l rrrr rr cc}
        \hline
        Dataset & \multicolumn{4}{c}{Video} & \multicolumn{2}{c}{Text} & Genre & Language \\
        & \# Video \ & \ Min. \  & \# Clip \  & Sec. \  & \# Sent \  & Len.  \ &  &  \\
        \hline \hline
        \multirow{2}{*}{VISA}      & \multirow{2}{*}{2K}  & \multirow{2}{*}{2.6} & \multirow{2}{*}{35K}  & \multirow{2}{*}{10.0} & \multirow{2}{*}{40K}  & \multirow{2}{*}{7.0}  & \  Film and & \multirow{2}{*}{Ja-En} \\
        &  &  & &  &  & & \ Television &  \\
        VATEX     & 25K & 0.2 & 25K  & 10.0 & 129K & 15.2 & Short Video         & En-Zh \\
        How2      & 13K & 1.2 & 186K & 5.8  & 186K & 20.6 & Short Video         & En-Pt \\
        BigVideo  & 156K & 3.8 & 4.5M & 8.0  & 4.5M & 22.8 & Short Video         & En-Zh \\
        \multirow{2}{*}{EVA}       & \multirow{2}{*}{2K} & \multirow{2}{*}{107.7}  & \multirow{2}{*}{1.3M} & \multirow{2}{*}{10.0} & \multirow{2}{*}{1.3M} & \multirow{2}{*}{7.3}  & \  Film and  & \ Ja-En \\
               &    & & & & & & \ Television & \ Zh-En \\
        TopicVD   & 256 & 66.8 & 122K & 8.4  & 122K & 12.5 & Documentary         & En-Zh \\
        \hline
    \end{tabular}
\end{table}

\section{Dataset}
\label{sec:data_cons}

\begin{figure}[t]
    \centering
    \includegraphics[scale=0.43]{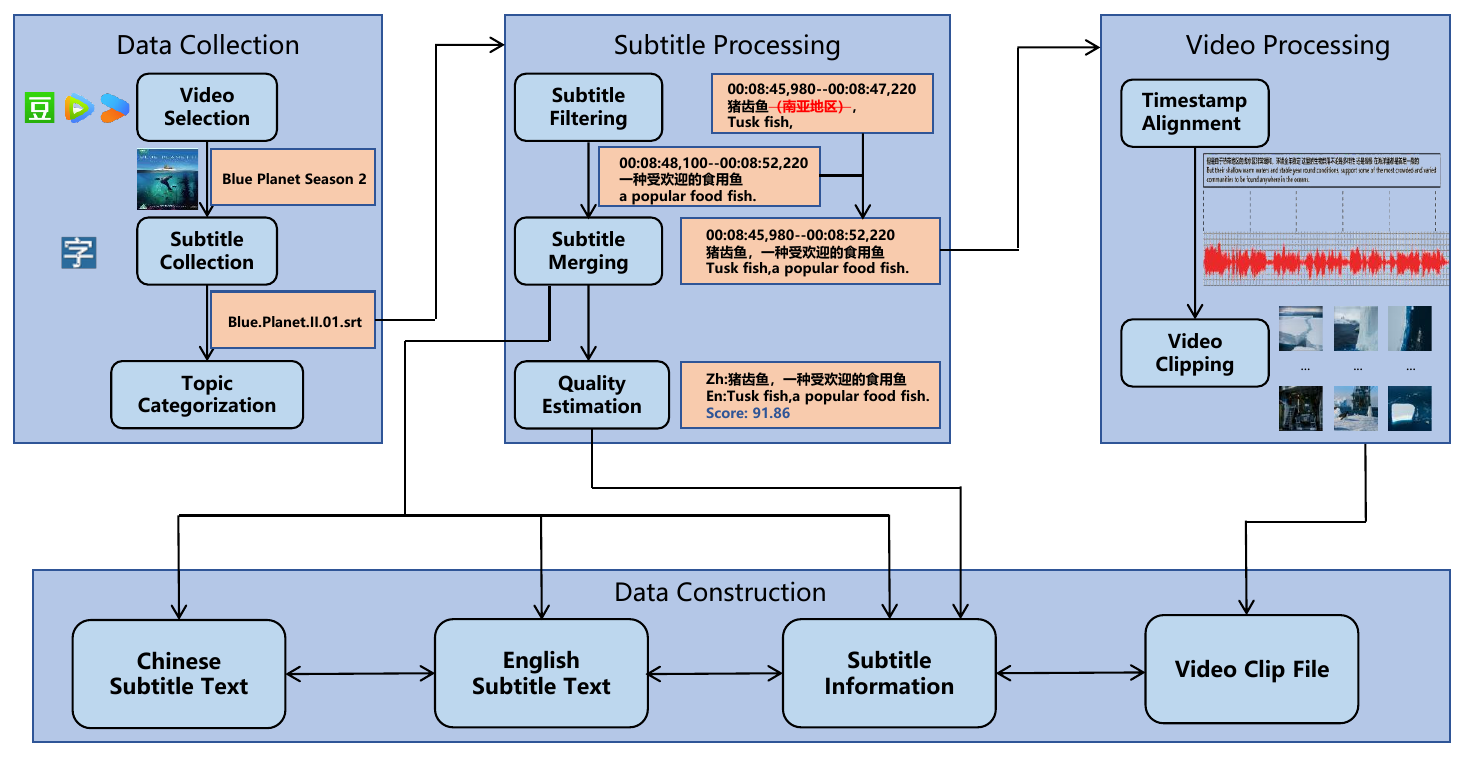}
    \caption{Construction process of the TopicVD documentary translation dataset}
    \label{fig:data_cons}
\end{figure}

The choice of documentaries as the basis for our video subtitle translation dataset is motivated by three key reasons.
First, documentary multimodal translation is a common translation scenario, yet no dedicated dataset currently exists for this task.
Second, documentaries span a wide range of topics, such as economics and nature, with significant domain-specific variations in content. This diversity makes them particularly suitable for studying domain adaptation in VMT.
Finally, documentaries are typically long in duration and maintain a high level of content consistency, making them well-suited for exploring VMT methods that leverage global video information.
Moreover, documentaries offer additional advantages, including easy data accessibility, a strong correlation between subtitle text and video content, and diverse, rich scenes. Therefore, we choose to construct the VMT dataset using documentaries and their subtitle information.
The dataset construction process is illustrated in Fig.~\ref{fig:data_cons}.

\subsection{Data Collection}
\label{sec:data_col}

The collected data consists of documentary videos along with corresponding Chinese and English subtitle files.
The specific collection steps are as follows:

\begin{itemize}
    \setlength{\itemsep}{0pt}
    \item[(1)] We select highly played and highly rated documentaries from video platforms such as Douban Movie~\footnote{\url{http://movie.douban.com}}, Tencent Video~\footnote{\url{https://v.qq.com}} and Youku Video~\footnote{\url{https://www.youku.com}}, ensuring that the dataset consists of high-quality and representative documentaries. 
    \item[(2)] We collect bilingual subtitle files from Subtitle Library~\footnote{\url{https://zimuku.vip/}} and SubHD~\footnote{\url{https://subhd.tv/}} for selected documentaries. These subtitle files contain text in both languages, such as Chinese and English.
    %
    \item[(3)] We assign a unique topic to each collected documentary. Since a documentary may cover multiple topics, we synthesize classification data from various video platforms and make final manual judgments on the topic of each documentary\footnote{
     This approach is evidently influenced by the topic distribution of publicly available documentaries, leading to a dataset that is predominantly composed of content related to Nature and History.
    }.
\end{itemize}

\subsection{Subtitle Processing}
\label{sec:subt_process}

The collected subtitle files are typically in SubRip Text (SRT) format, which includes code for controlling subtitle display formatting. To ensure readability, subtitle creators often split longer sentences into multiple segments. Thus, we preprocess the subtitle files, and the processing steps are as follows:

\begin{itemize}
    \setlength{\itemsep}{0pt}
    \item[(1)] We filter out the code from the text. To ensure correspondence between the two language texts, we further filter out special symbols that appear only in one language.
    \item[(2)] We use punctuation marks and corresponding time segment as cues to reassemble split segments into complete sentences. During the merging process, time segments must also be merged to ensure that the reassembled subtitles accurately align with their appearance in the documentary.
    \item[(3)] To ensure the quality of parallel subtitles, we compute the cosine similarity score of each subtitle pair using a multilingual sentence embedding model (e.g., MPNet~\cite{song2020mpnet}). This score serves as a quality estimation metric and can be used to filter out low-quality pairs across different experimental settings.
\end{itemize}


\subsection{Video Processing}
\label{sec:video_proc}

Based on the subtitle processing results from Section~\ref{sec:subt_process}, we further process the video information of the collected documentaries. 
The processing steps are as follows:

\begin{itemize}
    \setlength{\itemsep}{0pt}
    \item[(1)] We use open-source speech recognition engines (e.g., Whisper~\footnote{\url{https://github.com/openai/whisper}}) to verify the alignment between the spoken content in the video and the subtitle text, adjusting any inconsistencies. Manual verification ensures that the subtitles accurately correspond to their respective segments in the documentary.
    \item[(2)] Using the time segment information from the subtitle text, we employed FFmpeg~\cite{wales2024validation} to segment the video into multiple clips corresponding to the subtitles and recorded the position of each clip within the full video.
\end{itemize}

\subsection{Data Construction}
\label{sec:data_sum}

The bilingual subtitle texts processed in Section~\ref{sec:subt_process} and the video clips processed in Section~\ref{sec:video_proc} are then collected to construct the proposed topic-based datasets for video-guided multimodal machine translation of documentaries.
In addition to the bilingual subtitle text files and corresponding video clip files, we also recorded specific information for each video-subtitle pair. 
As shown in Table~\ref{tab:data_info_exp}, we documented the documentary title, topic, time segment, and the position of each pair within the documentary. 
Furthermore, we recorded the quality estimation scores of the bilingual subtitle text, as calculated in Section~\ref{sec:subt_process}. 
These scores not only reflect the semantic similarity between subtitle pairs but also serve as a quantitative basis for subsequent similarity-based filtering experiments.

\begin{table}[t]
\caption{Extracted specific information for each video-subtitle pair. ``Title'' and ``Topic'' represent the title and topic of the corresponding documentary, respectively. ``Start'' and ``End'' indicate the start and end timestamps of the video-subtitle pair, while ``Position'' denotes its relative placement within the documentary. ``Score'' refers to the quality estimation score of the bilingual subtitle text. }
\label{tab:data_info_exp}
\centering
\begin{tabular}{p{3.2cm} p{1.4cm} p{2cm} p{2cm} p{1.5cm}  p{1cm}}
\hline 
Title & Topic & Start & End & Position &  Score \\
\hline \hline
An Honest Liar.mp4  & Human & 00:00:02,960 & 00:00:09,800 & 1   & 92.93 \\
An Honest Liar.mp4  & Human & 00:00:12,040 & 00:00:15,040 & 2  & 92.75 \\
\multicolumn{6}{c}{\dots} \\  
An Honest Liar.mp4  & Human & 01:21:44,639 & 01:21:47,879 & 872 & 87.01 \\
An Honest Liar.mp4  & Human& 01:21:47,879 & 01:21:49,719 & 873  & 83.55 \\
\hline
\end{tabular}
\end{table}


We collected a total of 256 documentaries, amounting to 285 hours of video and 122,930 pairs of English-Chinese parallel subtitles. 
To mitigate the impact of random variation on experimental results, we randomly selected one to two documentaries per topic as the validation and test sets, while the remaining data were used for training. 
Note that for each topic, the training and test sets are drawn from the same topic to ensure topic-level relevance. However, the test set contains data from different documentaries than those in the training set, ensuring no direct content overlap.
Table~\ref{tab:dataset} presents a detailed statistical summary of the training, validation, and test sets.


\begin{table}[t]
    \caption{Data size statistics for training, validation, and test Sets by Topic in TopicVD (number of video-subtitle pairs / documentary videos)}
    \label{tab:dataset}
    \centering
    \begin{tabular}{l ccc c}
    \hline
    Topic \ \ \ \  &\  \ \ \ Train \  \ \ \  &  \ \ \ \ Valid \  \ \ \  &  \ \ \ \ Test \  \ \ \  & \ \ \ \  total \\
    \hline \hline
    Economy  & 5,080 / 7  & 1,483 / 1 & 1,904 / 1 & 8,467 / 9 \\
    Food   & 1,574 / 6 & 705 / 1 & 508 / 2 & 2,787 / 9 \\
    History  & 17,542 / 42 & 1,034 / 1  & 1,047 / 2 & 19,623 / 45        \\
    Figure   & 24,748 / 25 & 1,446  / 2 & 1,307 / 2 & 27,501 / 29       \\
    Military  & 2,162 / 3 & 1,036 / 4 & 1,138 / 1 & 4,336 / 8       \\
    Nature   & 26,015 / 90 & 1,474 / 5 & 1,482 / 8 & 28,971 / 103         \\
    Social   & 11,415 / 13 & 1,486 / 2 & 1,033 / 1 & 13,934  / 16       \\
    Technology   & 13,966  / 33  & 1,765 / 2 & 1,580 / 2 & 17,311  / 37       \\
    \hline
    Total   & 102,502 / 219 & 10,429  / 18  & 9,999  / 19    & 122,930  / 256       \\
\hline
\end{tabular}
\end{table}

\section{Model}
\label{sec:model}

\begin{figure}[t]
    \centering
    \includegraphics[scale=0.45]{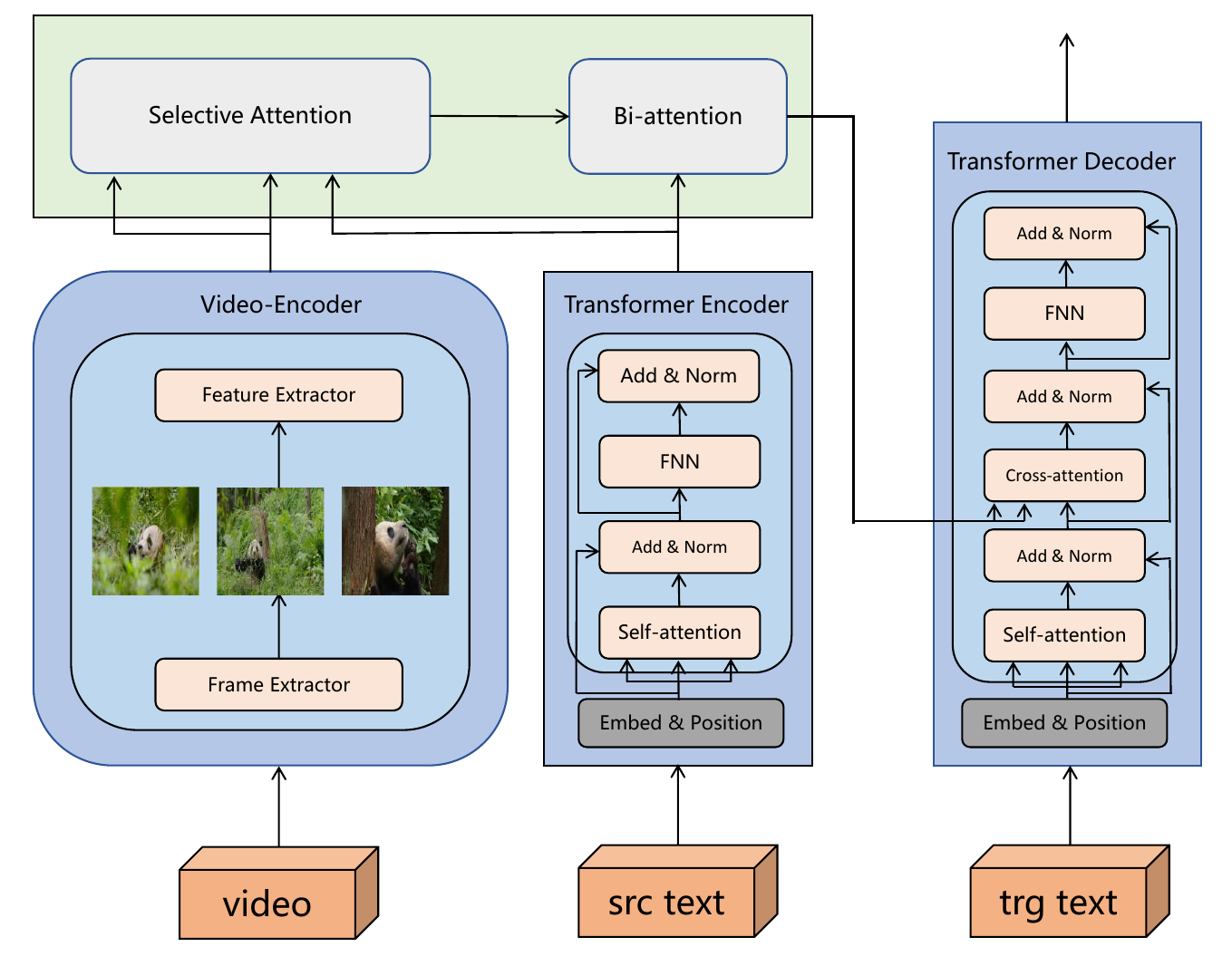}
    \caption{An overview of the proposed VMT model with cross-modal bidirectional attention}
    \label{fig:model}
\end{figure}

Inspired by the Selective Attention module~\cite{li2021vision} and the Bidirectional Attention module~\cite{tang2022multimodal} in IMT, we propose a VMT model incorporating a cross-modal bidirectional attention mechanism. 
Fig~\ref{fig:model} provides an overview of the model.

\subsection{Selective Attention}
\label{sec:sel_att}

To enhance the association efficiency between video features and text representations and achieve deep information fusion, this proposed model introduces a cross-modal selective attention mechanism.
Given a bilingual subtitle pair \( < (\mathbf{e}, \mathbf{f}), \mathbf{V} > \), where \( \mathbf{e} \) and \( \mathbf{f} \) represent the source and target texts, respectively, and \( \mathbf{V} \) denotes the corresponding video clip for the subtitle. 
The source text \( \mathbf{e} \) is encoded by a transformer encoder~\cite{vaswani2017attention}, producing the output text representation \( H^{\mathbf{e}} \). The video clip \( \mathbf{V} \) is then encoded by the I3D model~\cite{carreira2017quo}, with the output video feature denoted as \( H^{\mathbf{V}} \). 

It is important to note that, unlike encoding an image to obtain its semantic vector directly, encoding a video requires first extracting the most representative frames before encoding them.
Therefore, we employ a single-head attention network as a selective attention mechanism to establish associations between the text and the extracted video frames. 
In this attention mechanism, the text representation \( H^{\mathbf{e}} \) is used as the query, while the video feature \( H^{\mathbf{V}} \) serves as both the key and the value, enabling efficient filtering and feature extraction of video features based on the textual content.  
The selective attention output \( H^{\mathbf{V}}_{\mathrm{att}} \) is defined as:  
\[
H^{\mathbf{V}}_{\mathrm{att}} 
= \mathrm{Softmax}\!\biggl(\frac{H^{\mathbf{e}} (H^{\mathbf{V}}_{i})^\top}{\sqrt{d_{H^{\mathbf{e}}}}}\biggr) \, H^{\mathbf{V}}
\]
where \( d_{H^{\mathbf{e}}}\) denotes the dimensionality of  \( H^{\mathbf{e}} \), and \( H^{\mathbf{V}}_{i} \) represents the \(i\)th frame of \(H^{\mathbf{V}}\).

\subsection{Bi-Attention}
\label{sec:bi_att}

We further introduce a bidirectional attention mechanism to enhance the interaction between text and video features, thereby improving the feature representations of both.
The core idea of Bi-Attention module is to compute attention from two modal perspectives—text-to-video and video-to-text—through bidirectional interaction.

The text feature matrix is given by \( H^{\mathbf{e}} = (h_1, h_2, \dots, h_N) \), and the video feature matrix is \( H^{\mathbf{V}}_{\mathrm{att}} = (a_1, a_2, \dots, a_L) \), both of which represent the embedding sequences of the text and video, respectively. Here, \( N \) denotes the length of the text, and \( L \) represents the number of video regions after processing in Section~\ref{sec:sel_att}. 
In the Bi-Attention module, an alignment matrix \( S \in \mathbb{R}^{N \times L} \) is first computed as follows:  
\[
\mathbf{S}_{(n,l)} = g(h_n \cdot a_l) 
\]
where \( S_{(n,l)} \) represents the similarity between the \( n \)th word of the text and the \( l \)th region of the video, and \( g(\cdot) \) is a scalar function.  
Based on the alignment matrix \( \mathbf{S} \), enhanced text features \(\overline{h_{n}}\) and enhanced video features \( \overline{a_{l}} \)are obtained through weighted summation as follows.   
\begin{eqnarray}
w_{n}^{t2v} =  {\rm softmax}(S_{n:}) & \ \ \ & \overline{h_{n}} = h_{n} + \sum\limits_{l}w_{n}^{t2v}a_{l} \nonumber \\
w_{l}^{v2t} =  {\rm softmax}(S_{:l}) & \ \ \  & \overline{a_{l}} = a_{l} + \sum\limits_{i}w_{l}^{v2t}h_{n} \nonumber 
\end{eqnarray}
Finally, \( \overline{h_{n}} \) and \( \overline{a_{l}} \) are input into a transformer-based decoder, as shown in Fig.~\ref{fig:model}.  

\section{Experiments}
\label{sec:exp}

\subsection{Experiment Setting}
\label{sec:exp_set}

Our model follows the training structure of the Selective Attention model \cite{li2022vision}. 
The model comprises 4 encoder and 4 decoder layers, with a hidden size of 128 and a feed-forward network (FFN) filter size of 256. The multi-head self-attention mechanism employs 4 heads, and the dropout rate is set to 0.3. For training, we use the RAdam optimizer \cite{liu2019variance} with a learning rate of 0.005.
The batch size is set to 4,096 tokens, and the learning rate is set to 1e-4. Training is terminated if the cross-entropy loss on the validation set does not improve within 10 iterations. The I3D model is used to extract video features \cite{carreira2017quo}. 

As baselines, we used the open-source transformer-translator-pytorch project \footnote{\url{https://github.com/devjwsong/transformer-translator-pytorch}} as the text-only NMT model.
Additionally, we compared the proposed model with an IMT model that employs heuristic video frame extraction strategies. Specifically, frames are extracted at a fixed rate per second from the video clip and then filtered based on the structural similarity values~\cite{wang2004image} of neighboring frames, using a threshold of 0.5 to eliminate redundant frames. The remaining representative frames are then selected.
In addition, we compared the proposed model with the approach by Teramen et al.~\cite{teramen2024english}. In this approach, the start, middle, and end frames of each video clip are extracted and selected based on either text-text similarity or text-image similarity.
All the above models are trained on the dataset shown in Table~\ref{tab:dataset}, and the 4-gram BLEU score~\cite{papineni2002bleu} is used as the evaluation metric for both the baselines and the proposed method.
\subsection{Results}
\label{sec:result}

Table~\ref{tab:bleu} presents the experimental results.
First, as shown in Table~\ref{tab:bleu}, TopicVD achieves a minimum BLEU score of 19.14 across the dataset, demonstrating that its quality is sufficient for MMT research.
The proposed method achieves a BLEU score of 29.33.
Compared to the text-only NMT, the proposed method yields a significantly higher BLEU score.
Compared to the IMT method with heuristic frame extraction, the proposed method improves performance by approximately 2.6 BLEU.
Compared to the IMT model proposed by Teramen~\cite{teramen2024english}, the proposed method achieves an improvement of approximately 3.4 BLEU.

\begin{table}[t]
\centering
\caption{Comparison of different models on BLEU scores.}
\label{tab:bleu}
\begin{tabular}{l c}
\hline
Method & BLEU \\
\hline \hline
text-only NMT  & 19.14 \\ \hline
IMT using heuristic frame extraction & 26.71 \\
IMT using text-text similary-based frame extraction~\cite{teramen2024english}  & 25.73 \\
IMT using text-image similary-based frame extraction~\cite{teramen2024english}  & 25.94 \\ \hline
the proposed method   & \textbf{29.33} \\
\hline
\end{tabular}
\end{table}

\begin{figure}[t]
    \centering
    \includegraphics[scale=0.36]{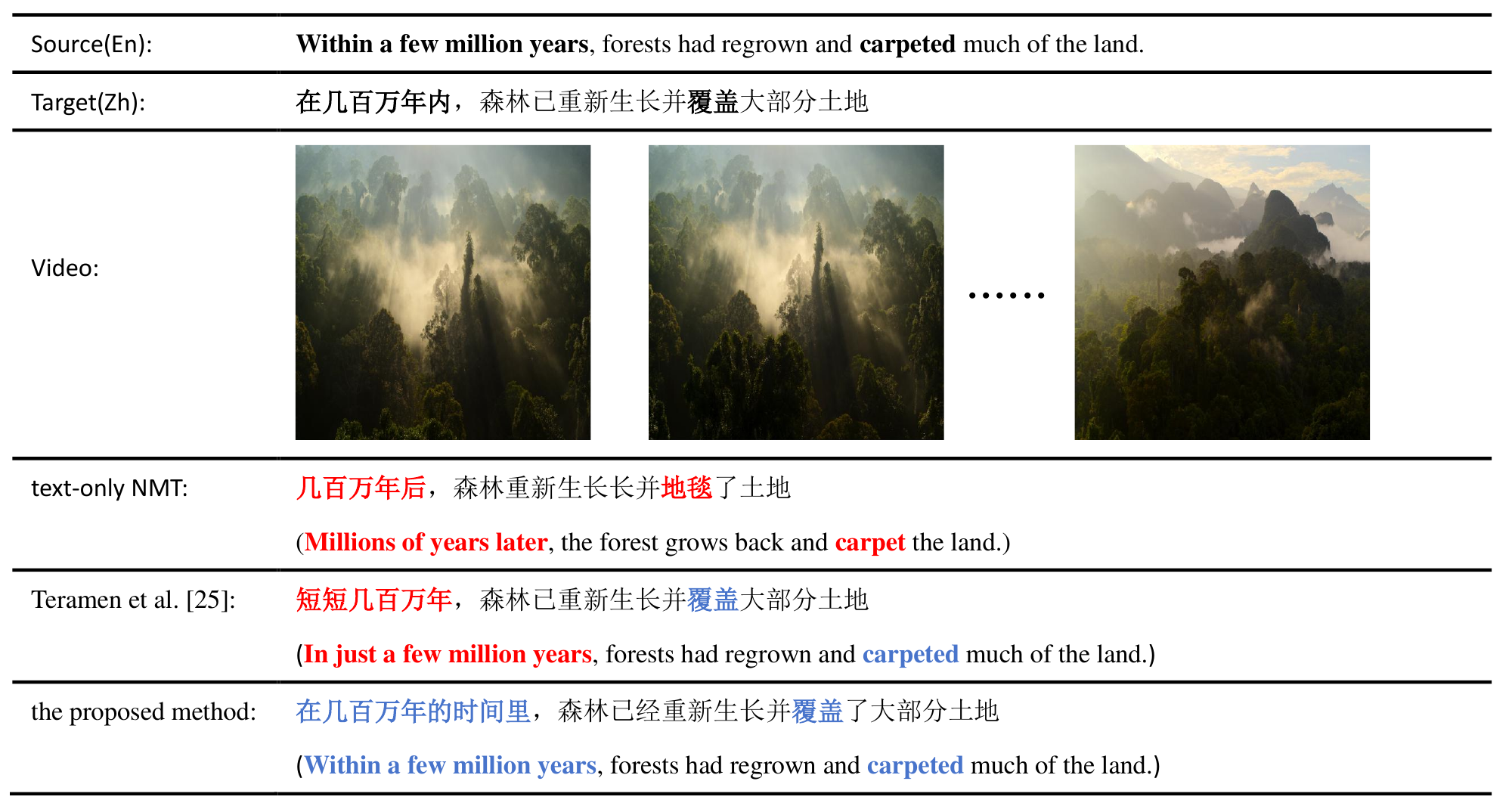}
    \caption{Example of correct translation by the proposed method}
    \label{fig:cor_exp}
\end{figure}

Fig.~\ref{fig:cor_exp} shows an example of correct translation by the proposed method.
In this example, the English phrase ``within a few million years'' fails to be translated by both the text-only NMT model and the model proposed by Teramen et al.\cite{teramen2024english}.
This is primarily because the cross-modal bidirectional attention mechanism enhances the performance of the text encoder.
Additionally, by leveraging visual information from the video clip related to ``carpeted'', the proposed method produces a correct translation, as does the IMT model proposed by Teramen et al.~\cite{teramen2024english}. 
In contrast, the text-only NMT model incorrectly translates ``carpeted'' into a Chinese noun meaning ``carpet''.
These findings highlight the effectiveness of incorporating visual information from videos in the VMT process for documentary translation.

\section{Discussion}
\label{sec:dis}


\subsection{In-Domain vs. Out-of-Domain Performance in VMT}
\label{sec:cross_topic}

To our knowledge, this study is the first to experimentally quantify the impact of in-domain and out-of-domain scenarios on VMT translation.
We conduct experiments to investigate the effect of varying data domains on the performance of the VMT model. 
From the TopicVD dataset, we select four topic subsets: History, Figure, Nature, and Technology.
These topics are chosen due to their large dataset sizes and the availability of relevant resources.
Consequently, we modify the training set in four distinct ways and train the VMT model proposed in Section~\ref{sec:model}.
Finally, we evaluate the model using the corresponding test sets for each selected topic.

The experimental details are as follows:
\begin{description}
    \setlength{\itemsep}{0pt}
    \item[Full dataset] We utilize the complete training data from all eight topics and evaluate the trained model using the corresponding test sets for each selected topic.
    \item[In-domain dataset] We train the model on a training set that corresponds to the same topic as its respective test set.  
    Clearly, for each test set of a selected topic, we train a separate in-domain VMT model.
    \item[Out-of-domain dataset (full)] We train the model using the training data from seven topics, excluding the topic of the test set.  
    Clearly, for each distinct test topic, we train a separate out-of-domain VMT model.
    \item[Out-of-domain dataset (sampled)] To ensure comparability with the in-domain dataset, we randomly sample a subset from the training data of seven topics, excluding the test set’s topic, ensuring the same data volume as the in-domain dataset of the test set.  
    For each distinct test topic, we train a separate out-of-domain VMT model.
\end{description}

\begin{table}[t]
    \caption{Translation performance of each experimental scenario across each test set (BLEU score)}
    \label{tab:combined_results}
    \centering
    \begin{tabular}{lcccc}
        \hline
        & History & Figure & Nature & Technology \\
        \hline \hline
        Full dataset & 24.13 & 25.15 & 32.70 & 31.16 \\
        In-domain dataset & 15.78 & 17.45 & 20.16 & 15.51 \\
        Out-of-domain dataset (full) & 22.14 & 24.76 & 28.94 & 29.66 \\
        Out-of-domain dataset (sampled) & 10.28 & 11.63 & 16.18 & 10.51 \\
        \hline
    \end{tabular}
\end{table}

Table~\ref{tab:combined_results} presents the BLEU scores obtained from the four aforementioned experimental scenarios.
For the selected four test topics, the full dataset achieved the best translation performance.
When comparing the results of the in-domain and out-of-domain (sampled) models, the in-domain model demonstrated significantly better translation performance, highlighting the effectiveness of in-domain data.
However, when comparing the in-domain results with those of the out-of-domain (full) model, the latter achieved a higher BLEU score.
This may be due to the fact that a sufficient amount of data can partially mitigate translation issues caused by out-of-domain data.
The comparison between the full dataset and the out-of-domain (full) results further supports these findings.
Specifically, for the Figure topic results, the out-of-domain (full) model showed the smallest decline compared to the full dataset. This is because, unlike the History, Nature, and Technology topics, the Figure topic covers a broader range of content.
As a result, in this case, the influence of in-domain data is more easily replaced by out-of-domain data.

\subsection{Impact of Data Augmentation on In-Domain Performance}
\label{sec:data_aug}

We further investigate the impact of data volume on the translation performance of the in-domain model.
To achieve this, we manually augment the training datasets for the Nature and Technology topics.
Specifically, we added 31 documentaries, comprising a total of 6,433 video-subtitle pairs, to the Nature training set, and 10 documentaries, totaling 2,688 video-subtitle pairs, to the Technology training set.
The experimental results are shown in Table~\ref{table:data_aug}.
The in-domain dataset refers to the VMT model trained on a training set that shares the same topic as the corresponding test set.
Specifically, the Nature training set contains 26,015 video-subtitle pairs, while the Technology training set comprises 13,966 video-subtitle pairs.
In contrast, the in-domain (augmented) dataset refers to the VMT model trained on the augmented training set.
Specifically, the augmented Nature training set contains 32,488 video-subtitle pairs, while the augmented Technology training set comprises 16,654 video-subtitle pairs.

As shown in the table, increasing the volume of in-domain data significantly improves the translation performance of the VMT model, with a greater effect than an equivalent increase in out-of-domain data.
Among all topics, the Nature topic exhibits the highest sensitivity to data augmentation, suggesting that its translation performance is highly dependent on in-domain data. 
A detailed analysis of the data reveals that the newly added Nature dataset contains a documentary closely related to the test set content, which significantly enhances translation performance.
This finding suggests a relatively straightforward approach to the domain adaptation problem in documentary translation—targeted augmentation with test-related data can effectively improve translation performance.

\begin{table}[t]
\centering
\caption{Translation performance of augmented train set across each test set (BLEU score)}
\label{table:data_aug}
\begin{tabular}{l c c}
\hline
 & Nature  & Technology \\
\hline \hline
 In-domain dataset & 20.16  &  15.51 \\
 In-domain dataset (augmented) & 30.60  & 18.62 \\
\hline
 Out-of-domain dataset  & 16.18  &  10.51 \\
 Out-of-domain dataset (augmented) & 17.35  & 13.54 \\
\hline
\end{tabular}
\end{table}

\subsection{Impact of Contextual Information on the VMT Model}
\label{sec:global_eva}

We further explore the impact of contextual video information in the VMT task. This experiment is conducted exclusively on the TopicVD dataset, where the positional information of each video-subtitle pair is preserved.
Here, we introduce a contextual video information filtering strategy based on text-text similarity. 
For a given subtitle text to be translated, we compute its similarity with other subtitle texts from the same documentary and select the video clip corresponding to the most similar subtitle as its contextual video information.
Table~\ref{tab:global_fusion} presents the VMT translation performance under three conditions: using only the video clip corresponding to the subtitle text, using three clips, and using ten clips.
The experimental results show that translation performance improves as the number of video clips increases from one to three to ten. This suggests that a moderate increase in video clips provides more contextual and semantic information to the model in the VMT task, further confirming the crucial role of contextual video information in VMT.

\begin{table}[t]
    \centering
    \caption{Translation performance with different contextual information}
    \label{tab:global_fusion}
    \begin{tabular}{lc}
        \hline
        \textbf{Method} & \textbf{BLEU} \\
        \hline \hline
        VMT with a single video clip & 29.33 \\
        VMT with three video clips & 29.95 \\
        VMT with ten video clips & \textbf{30.22} \\
        \hline
    \end{tabular}
\end{table}

\section{Conclusion}
\label{sec:concl}

In this study, we introduced TopicVD, a topic-based dataset for video-supported multimodal machine translation of documentaries, to advance research in documentary translation using multimodal machine translation.
We propose an MMT model based on a cross-modal bidirectional attention module and validate the effectiveness of both the dataset and the model through experiments.
Extensive experiments on the TopicVD dataset show that the performance of the MMT model significantly declines in out-of-domain scenarios, highlighting the need for effective domain adaptation methods.
Additionally, experiments demonstrate that global context can substantially improve translation performance.
These findings underscore the importance of topic-specific data organization and contextual modeling in complex video-guided machine translation tasks.
As a future direction, we will focus on TopicVD to investigate domain adaptation methods for documentary VMT and explore how contextual information can optimize documentary VMT performance. 
In addition, we will expand the TopicVD dataset and diversify its topics to enhance its practical value.

\bibliographystyle{splncs04}
\bibliography{custom}

\end{document}